\documentclass[10pt,twocolumn,letterpaper]{article}

\usepackage{cvpr}              
\definecolor{cvprblue}{rgb}{0.21,0.49,0.74}
\usepackage[pagebackref,breaklinks,colorlinks,allcolors=cvprblue]{hyperref}
\usepackage{makecell}
\usepackage{textcomp}
\usepackage{stfloats}
\usepackage{url}
\usepackage{verbatim}
\usepackage{graphicx}
\usepackage{hyperref}       
\usepackage{amssymb}
\usepackage{multirow}
\usepackage{tabularx}
\usepackage{wrapfig}
\usepackage{multirow}
\usepackage{makecell}
\usepackage{changepage}
\usepackage{setspace}
\usepackage{pifont}
\usepackage{xspace}  
\usepackage{booktabs}
\usepackage{xcolor}
\usepackage{subcaption}
\usepackage{threeparttable}
\def\confName{CVPR}
\def\confYear{2026}

\title{ODOV: Benchmark the Open-Domain Open-Vocabulary Object Detection}

\author{
	Yupeng Zhang\textsuperscript{\rm 1,2}\hspace{1em}
	Ruize Han\textsuperscript{\rm 3}\thanks{Corresponding author.}\hspace{1em}
	Fangnan Zhou\textsuperscript{\rm 1}\hspace{1em}
	Wei Feng\textsuperscript{\rm 1,2}\hspace{1em}
	Liang Wan\textsuperscript{\rm 1,2}\hspace{1em}\\
	\textsuperscript{\rm 1}College of Intelligence and Computing, Tianjin University.\\
    \textsuperscript{\rm 2}Key Research Center for Surface Monitoring and Analysis of Relics, State Administration of Cultural Heritage.\\
	\textsuperscript{\rm 3}Faculty of Computer Science and Artificial Intelligence, Shenzhen University of Advanced Technology.\\
{\tt\small \{zhangyupeng, zhoufangnan, wfeng, lwan\}@tju.edu.cn, hanruize@suat-sz.edu.cn}
}

\begin{document}
\maketitle
\begin{abstract}
Existing studies typically investigate domain shift and category shift as independent problems, however, in real-world scenarios, the two types of shifts often occur simultaneously and interact, leading to significant degradation in detection performance. 
To address this, we propose and systematically study a novel problem—Open-Domain Open-Vocabulary (ODOV) object detection—which aims to evaluate a model’s ability to adapt to the compound domain and category shifts in \textbf{real-world environments}.
We construct a new benchmark, OD-LVIS, which contains 46,949 images spanning 15 diverse real-world scenarios and 1,203 categories, for assessing object detection performance.
Furthermore, we propose a novel ODOV detection baseline that fully leverages VLM's powerful multi-modal alignment capabilities and introduces two key mechanisms to enhance both category and domain generalization. One is the Domain-Agnostic Category Prompt (DAPmt), which strengthens category semantics while attenuating domain representations, enabling pure category representation. 
The other is the Domain Projection and Grafting (DP\&G) module, which incorporates domain-specific features from input images, allowing the model to dynamically generalize across diverse open domains. 
These two components enable the model to maintain effective detection performance under simultaneous category and domain variations in real-world scenarios.
We provide extensive benchmark evaluations for the proposed ODOV detection task and report experimental results. These results validate the soundness of the ODOV task, the practicality of the \href{https://2899253375.github.io/OD-LVIS/}{OD-LVIS} dataset, and the superiority of the method.

\end{abstract}

\section{Introduction}
Object detection is a fundamental task in computer vision, which aims at locating and identifying objects within images. 
Recent years have witnessed the rapid development of object detection, which, however, is still with unsolved problems for real-world applications. 
On one hand, due to the labor-intensive and costly manual annotation process for bounding boxes, the annotated object categories in the object detection dataset (for training) are limited, which are certainly smaller than the category vocabulary in the real world (\textbf{category shift}).
On the other hand, most existing works have primarily focused on handling clear natural images. 
However, in many dynamically evolving real-world scenarios, such as autonomous driving, video surveillance, and internet search, the acquired images are often neither high-quality nor high-resolution.
Instead, they may be plagued by various types of degradation or style differences (\textbf{domain shift}). 
For instance, the images captured in rainy or foggy weather (from the scene), with low resolution or various noises (during the imaging), or with different artistic styles (Internet data), \etc.

\begin{figure}[t!] 
	\centering
	\includegraphics[width=0.88\linewidth]{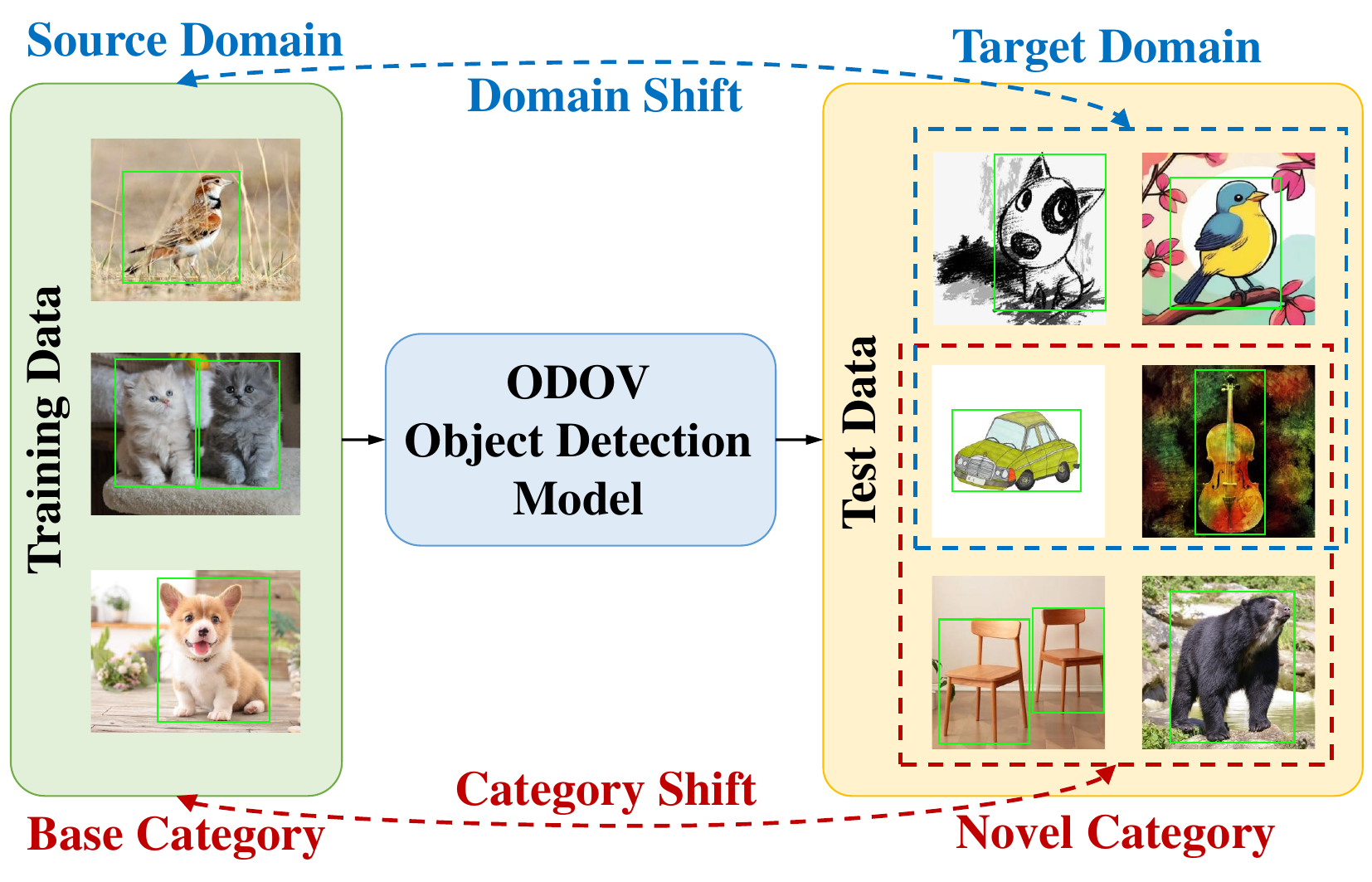}\vspace{-10pt}
	\caption{Illustration of the ODOV object detection model, showing the simultaneous occurrence of category and domain shifts. Blue dashed lines indicate domain shifts, primarily reflected in stylistic variations—for example, the training data comprise real-world photographs, while the test data may include cartoon or watercolor-style images. Red dashed lines indicate category shifts, characterized by changes in category distribution, such as new categories (\eg, chairs, cars) appearing in the testing stage.}\vspace{-10pt}
	\label{fig:motivation} \vspace{-10pt}
\end{figure}

We find that current research on object detection (OD) in open scenes has noticed the above two problems, however, they are often studied in isolation. 
Such as the one-shot OD~\cite{hsieh2019one,chen2021adaptive,zhao2022semantic,yang2022balanced}, open-world OD~\cite{wang2023detecting,ma2022rethinking,mullappilly2024semi}, and the most recent open-vocabulary OD~\cite{gu2021open,wu2023clipself} are designed for the open-category detection problem.
Besides, for the open-domain setting, domain adaption, domain generalization have been used for OD task~\cite{fan2023towards,lin2021domain,zhang2022gated,wu2022single,vidit2023clip,rao2023srcd,wu2024g}, and some image restoration methods are also applied to the downstream OD task~\cite{wang2022togethernet,wang2024joint,wang2024degradation}. 
Actually, the \textit{compound category and domain shifts are more common} in real-world scenes. 
For example, from indoor (training) to outdoor (testing) scenes, object categories often change significantly, while the visual domain of images also shifts, such as under different lighting, weather conditions, \etc.

In this work, we propose a novel problem, namely open-domain open-vocabulary (ODOV) object detection to address the compound generalization of categories and domains, as shown in Fig.~\ref{fig:motivation}.
Specifically, during training, we learn the detection model \textit{only on the base categories with the natural domain}. In testing, we expect the model can detect and classify \textit{the objects with unseen categories and domains}. 
Clearly, this problem is more practical yet challenging, it faces two main new difficulties.

\begin{itemize}
\item \textbf{Challenge distinction}: Category generalization and domain generalization usually rely on distinct mechanisms. Category generalization emphasizes enhancing a model’s ability to recognize diverse objects, whereas domain generalization focuses on adapting to shifts in visual distributions, such as variations in artistic style or environmental conditions. 
Improper integration of these two mechanisms may undermine both category recognition and domain generalization,
severely degrading overall detection performance. Thus, striking a balance between category and domain generalization constitutes the central challenge and research value of ODOV.

\item \textbf{Benchmark shortage}: Existing large-scale detection datasets (\eg, LVIS) cover only a single domain (natural) images and fail to capture the complexity and diversity of real-world scenarios. In contrast, open-domain detection datasets (\eg, BDD100K~\cite{yu2020bdd100k}) include multiple visual domains but remain restricted to a few categories such as pedestrians and vehicles. As a result, current benchmarks either offer rich category diversity without multi-domain coverage or provide multi-domain variation with limited categories, and thus never simultaneously address the research needs of both domain and category shift.
\end{itemize}

Considering these two difficulties, we construct a new benchmark and establish the first unified framework for ODOV object detection, aiming to encourage attention to and exploration of solutions to this critical setting.
\textit{With respect to the benchmark}, we introduce a new evaluation dataset to systematically assess algorithms on the ODOV object detection task. Specifically, we design it based on the LVIS validation set and name it the Open-Domain LVIS benchmark (OD-LVIS). OD-LVIS contains 46,949 images, covering the same 1,203 categories as LVIS (including both base and novel categories), and is randomly distributed across 15 diverse and complex real-world domains, thereby more faithfully simulating continuously changing open environments. 
For models trained on LVIS base categories, OD-LVIS further incorporates rare (novel) categories and more challenging domain environments, thereby providing a more comprehensive and practical benchmark for ODOV object detection.
\textit{With respect to the method}, unlike previous approaches that leverage large vision-language models (VLMs) such as CLIP~\cite{radford2021learning} and ALIGN~\cite{jia2021scaling} to address either domain generalization~\cite{shu2023clipood} or open-vocabulary detection~\cite{gu2021open,zhong2022regionclip,minderer2022simple,bangalath2022bridging,kuo2022f,xu2023exploring,li2023distilling,wu2023clipself,wang2025ov}, 
we propose a novel strategy, designed to fully exploit the cross-modal alignment capabilities of VLMs and uncover their generalization potential across both semantic and domain dimensions. By dynamically integrating domain representations with category descriptions, we construct adaptive prompts that align with the semantic and domain characteristics of the input image, effectively addressing the intertwined challenges of category shift and domain shift in real-world scenarios.

Specifically, we leverage CLIP, a model trained on massive categories and domains with powerful representational capacity. 
To enhance the robustness of text embeddings under category and domain shifts, we guide a large language model with instructions to generate descriptions that highlight distinctive category attributes while downplaying domain-specific information. These serve as \underline{D}omain-\underline{A}gnostic Category \underline{P}ro\underline{m}p\underline{t}s (DAPmt), mitigating the ambiguity  (\eg, bat may refer either to a flying animal or a baseball bat) and domain constraints introduced by generic text prompts.
We further design a \underline{D}omain \underline{P}rojection and \underline{G}rafting (DP\&G) module that extracts domain-specific embeddings from input images and fuses them with DAPmt to generate \textbf{domain-customized category embeddings}. This mechanism enables the model to dynamically construct customized category text embeddings for each image, significantly enhancing its domain generalization capability in ODOV object detection.
On OD-LVIS, we evaluate five backbones—CLIP RN50×16, CLIPSelf ViT-B/16 and ViT-L/14, and DeCLIP ViT-B/16 and ViT-L/14—and obtain gains of 1.8\%, 2.1\%, 1.5\%, 1.8\%, and 1.4\% in $AP$, respectively. Moreover, our method consistently achieves the best overall performance compared with existing approaches.







\section{Related Work}
\label{sec:Related work}

\textbf{Open-vocabulary object detection (OVD)}~\cite{zareian2021open} aims to detect objects from novel categories unseen during training. Leveraging the zero-shot capabilities of vision-language models (VLMs), recent advancements in OVD have emerged. Works like~\cite{kim2023contrastive,kim2023detection,kim2023region,wu2023cora,song2023prompt} use region-aware training to integrate image-text pairs, improving classification, especially for novel categories. Studies such as~\cite{zhou2022detecting,zhong2022regionclip,jeong2024proxydet,ma2024codet,kaul2023multi} use large-scale image-text data, pre-trained VLMs, or pseudo-labels to predict novel categories and fine-tune the model with both pseudo and base labels. Other methods~\cite{gu2021open,pham2024lp,bangalath2022bridging,li2023distilling,du2022learning} employ knowledge distillation from VLMs to achieve OVD, transferring knowledge while enhancing localization. Some approaches build detectors on frozen VLMs~\cite{kuo2022f,minderer2022simple,wu2023clipself,wang2025declip,wang2025ov}, avoiding knowledge loss in fine-tuning and maximizing generalization. In this work, we use frozen VLMs as encoders, exploring their potential for category and domain generalization through prompt adjustment, revealing their strengths in open-domain and open-vocabulary object detection.

\textbf{Domain Generalization (DG) based object detection} aims to train a detector on multiple source domains to generalize to unseen target domains. Early work~\cite{lin2021domain} used feature disentanglement for cross-domain generalization, followed by the Gated Disentangling Network~\cite{zhang2022gated}, which activates feature channels for domain-invariant aspects. However, these methods rely on multiple domains and domain labels. Single Domain Generalized (SDG) object detection~\cite{wu2022single} addresses training with only one source domain. CDSD~\cite{wu2022single} separates domain-invariant from domain-specific representations, CLIP the Gap~\cite{vidit2023clip} uses pre-trained VLM, SRCD~\cite{rao2023srcd} reduces spurious correlations, and G-NAS~\cite{wu2024g} introduces a generalization loss to prevent Neural Architecture Search (NAS) overfitting. In this work, we adopt the SDG setting, using single-source domain data for training and evaluating multiple open domains. Our benchmark includes 15 diverse open domains, providing a more complex testbed than prior works.

\textbf{Prompt learning for VLM adaptation.} VLMs~\cite{radford2021learning, jia2021scaling, zhai2022lit, yao2021filip, yuan2021florence, sun2023eva} bridge image and text effectively. Pretrained on vast image-text pairs, models like CLIP~\cite{radford2021learning} excel in open-scene recognition. However, adapting them to specific tasks with limited data is challenging. Text prompts guide VLMs, but even advanced prompt learning methods~\cite{zhou2022learning, zhou2022conditional, khattak2023maple} still require training data. The recently proposed Test-time Prompt Tuning (TPT)~\cite{shu2022test} optimizes prompts by minimizing entropy through confidence-based selection, ensuring consistent predictions across different augmented views of each test sample. However, it remains inadequate when confronted with distribution shifts.
Our work explores VLM adaptation for open-domain and open-vocabulary settings at test time. By fusing style features with category descriptors, we dynamically create task-relevant embeddings, enhancing generalization across new categories and domains. 
Unlike traditional test-time prompt tuning (\eg, TPT), our approach does not require continuous adaptation on a single-style dataset. Instead, it directly addresses randomly varying distribution shifts, thereby substantially enhancing the model’s cross-domain generalization across both categories and domains.

\section{ODOV Object Detection Benchmark}
\label{sec:Benchmark}

\subsection{Motivation}
As shown in Table~\ref{tab:dataset}, we summarize existing object detection datasets as follows.
1) \textit{{General-category detection}}: Clipart~\cite{inoue2018cross}, Watercolor~\cite{inoue2018cross}, Comic~\cite{inoue2018cross}, and MSOSB~\cite{zhang2024rethinking}, with artistic styles; MS COCO~\cite{lin2014microsoft}, Objects365~\cite{shao2019objects365}, ODinW~\cite{li2022elevater} and LVIS~\cite{gupta2019lvis} consisting of common images with LVIS offering a more extensive set of categories often used for OV tasks.
2) \textit{Specific-category detection}: WIDER FACE~\cite{yang2016wider}, a dataset representing common photographic scenes, focused solely on the face category. {Besides, pedestrian detection and vehicle detection are also very popular with a series of datasets~\cite{zhang2017citypersons, dollar2009pedestrian, 9247499}, \etc.} 
3) \textit{Traffic scene detection}: Cityscapes~\cite{cordts2016cityscapes}, BDD100K~\cite{yu2020bdd100k}, Foggy Cityscapes~\cite{sakaridis2018semantic}, UFDD~\cite{nada2018pushing}, RTTS~\cite{li2018benchmarking}, and Sim10K~\cite{johnson2016driving} contain unique weather conditions to test the domain generalization performance of models, but primarily focus on a limited set of objects common in traffic scenarios.  
Overall, the above \textbf{existing datasets do not meet the requirement of containing simultaneous open-domain and open-category scenes}.

\begin{table}[t]
	\centering
    \scriptsize
	\caption{Comparison of OD-LVIS and other detection datasets.} \vspace{-10pt}  
	\label{tab:dataset}
	\begin{tabular}{l@{\hskip 4pt}c@{\hskip 4pt}c@{\hskip 4pt}c@{\hskip 4pt}c@{\hskip 4pt}c@{\hskip 4pt}c@{\hskip 4pt}c@{\hskip 4pt}c@{\hskip 4pt}}
		\toprule
		Name    &\#Year   &\#Image   &\#Category   &\#Domain  &\#Domain Type  \\
		\midrule
		MS COCO (val)  & 2014 &5,000  &80 &1 &Normal   \\
		Cityscapes  & 2016 &3,475  &8 &1 &Weather   \\
		Sim10K & 2016 &10,000  &1 &1 & Weather  \\
		WIDER FACE &2016 &32,000  &1 &1 &Normal   \\
		Foggy Cityscapes  &2018 &3,475  &8 &1 &Weather    \\
		Clipart  & 2018   &1,000   &20  &1 & Art   \\
		Watercolor  & 2018  &1,905   &6  &1  & Art   \\
		Comic & 2018 &1,905  &6 &1 & Art    \\
		UFDD &2018 &884  &1 &1 &Normal     \\
		RTTS  &2018 &9,109  &5 &1 & Weather  \\
        Objects365 &2019 & 100,000  & 365  & 1  & Normal\\
		LVIS (val) & 2019 &19,809  &1,203 &1 &Normal   \\
		BDD100K &2020  &41,986  &10 &12 & Weather  \\
        ODinW35 &2022  & 20,000  & 314  & 1  & Normal\\
		MSOSB  & 2024 & 76,146  & 80  & 5 & Art   \\  %
		\textbf{OD-LVIS} &2025 &\textbf{46,949}  &\textbf{1,203} &\textbf{15} &\textbf{\makecell{Art, Weather, \\Noise, Blur...}} \\
		\bottomrule
	\end{tabular} \vspace{-15pt}
\end{table}

\begin{figure}[t!] \vspace{-5pt}
	\centering
	\includegraphics[width=0.88\linewidth]{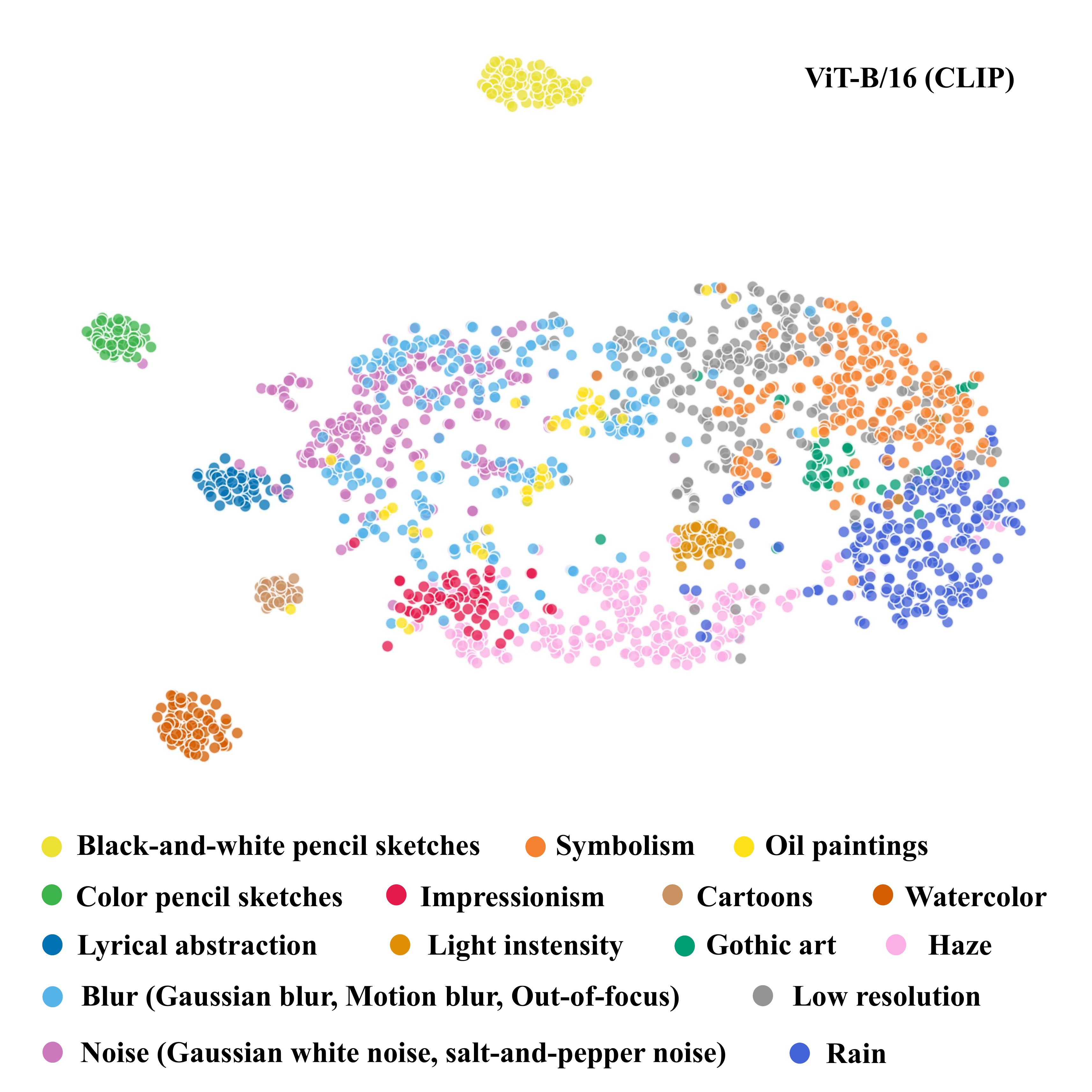}  \vspace{-10pt}
	\caption{The t-SNE visualization of feature-level style statistics from the CLIP image encoder (ViT-B/16) outputs for `apple' samples. The samples cluster by domain characteristics.}  
    \vspace{-20pt} 
	\label{fig:distri}
\end{figure}

This way, we build OD-LVIS, a dedicated evaluation benchmark designed specifically for ODOV object detection, encompassing a diverse range of categories and complex real-world scenarios. 
Specifically, we select the LVIS as our basic dataset.
On the one hand, we retain all object categories from LVIS (including both base and novel categories) to ensure the benchmark’s category diversity.
On the other hand, to enhance the domain diversity, we extend the data by considering two aspects, \ie, the \textit{image styles} and the \textit{imaging conditions}. Specifically, for the former, we collect nine distinct styles, \ie, black-and-white pencil sketches, color pencil sketches, oil paintings, cartoons, watercolors, symbolism, impressionism, gothic art, and lyrical abstraction.
For the latter, we also consider six imaging conditions, \ie, rain, haze, illumination variations, low resolution, noise (Gaussian white noise and salt-and-pepper noise), and blur (Gaussian blur, motion blur, and out-of-focus).
As a result, OD-LVIS comprises 46,949 images across 15 different real-world domains, shares categories with LVIS, and adheres to its annotation guidelines, establishing a standardized benchmark for ODOV object detection evaluation. As shown in Fig.~\ref{fig:distri}, we extract features of the apple category samples using the CLIP image encoder (ViT-B/16) and visualize them with t-SNE, clearly showing that the samples cluster according to their domain characteristics. Finally, since OD-LVIS shares categories with LVIS, it can be combined with LVIS as training data to further evaluate the domain generalization ability of object detection models across 15 different scenarios.

\subsection{Preliminary Experiments on OD-LVIS}
We investigate the impact of OD-LVIS on existing detector. 
Take the haze and noise for example, we evaluate the SOTA OVD method F-ViT (DeCLIP ViT-B/16)~\cite{wang2025declip} in Table~\ref{tab:degra}, where $m$ and $k$ control the levels of haze and noise, respectively (larger values indicate more severe shifts). 
Results show that as domain shifts increase, model performance declines significantly, demonstrating the necessity of the studying on ODOV detection and OD-LVIS benchmark.

\textbf{Please refer to the supplementary material for details on Data Generation, Cleaning, and Annotation.}

\vspace{-10pt}
\begin{table}[htpb]  
        \caption{Results on images with varying degrees of degradation.}\vspace{-10pt}
	\label{tab:degra}
	\centering
	\scriptsize
	\begin{tabular}{lcc}
    	\toprule
    	Data    & Degree / AP (\%)  & Degree / AP (\%)  \\\hline
        Source &- / 26.5     &- / 26.5  \\
        Haze   & $m$=0.05 / 24.5 &  $m$=0.08 / 20.7  \\
        Noise   & $k$=0.04 / 13.7 & $k$=0.06 / 12.3  \\
    	\bottomrule    
	\end{tabular} \vspace{-10pt}
\end{table}


\section{The Proposed Baseline Method} 
\label{sec:Method}

\subsection{Problem Formulation}
We first provide the problem formulation of the proposed ODOV object detection problem.
During the training stage, we use the data from the \textit{single source domain} (\ie, the natural image domain). Specifically, the training image dataset is notated as $\mathcal{D}^\mathrm{train} = \{\mathbf{I}_i^\mathrm{train}, L_i\}_{i=1}^{N}$, where $N$ is the number of training images, $\mathbf{I}_i$ denotes an image from the source domain with the detection latel $L_i$ for it.
The label $L_i$ is composed of $\{\mathbf{b}_i, \mathbf{c}_i\}$, in which $\mathbf{b}_i$ indicates all the annotated object bounding boxes in $\mathbf{I}_i$, and the corresponding object categories are stored in $\mathbf{c}_i$.
Note that, all the (annotated) object categories contained in $L_i$ of the training set $\mathcal{D}^\mathrm{train}$ are from the \textit{base category} set, \ie, $\mathcal{C}^\mathrm{base}$.

During the testing stage, the ODOV object detection task requires the model to be applied under open-domain conditions as well as open-vocabulary conditions. Specifically, the testing images are from hybrid open domains (\eg, with various image styles), which are denoted as $\mathcal{D}^\mathrm{test} = \{\mathbf{I}_j^\mathrm{test}\}_{j=1}^{M}$ and $M$ is the dataset scale.
For each test image $\mathbf{I}_j^\mathrm{test}$, the desired output is the predicted object bounding boxes with corresponding categories, \ie, $\{\mathbf{b}_j, \mathbf{c}_j\}$.
Note that, following the open-vocabulary detection setting, the predicted objects contain both the \textit{base categories} $\mathcal{C}^\mathrm{base}$ (appearing in training) and the \textit{novel categories} $\mathcal{C}^\mathrm{novel}$ (unseen during training), which are combined as the open-vocabulary category set, \ie, $\mathcal{C}^\mathrm{open} = \mathcal{C}^\mathrm{base} + \mathcal{C}^\mathrm{novel}$.





\subsection{Overview of The Method} 
\begin{figure*}[t!]
	\centering
	\includegraphics[width=0.9\linewidth]{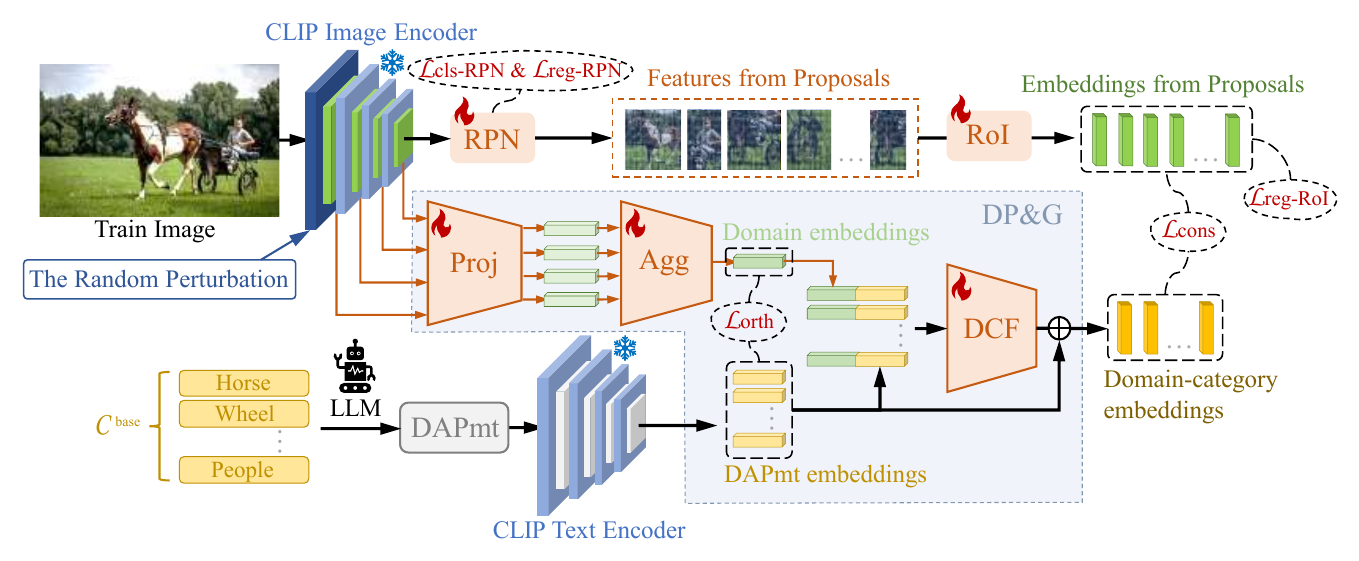}  \vspace{-12pt}
	\caption{Overview of DVtor. We use the generalization ability of pre-trained VLM (CLIP) by decoupling domain-aware representations and combining them with domain-agnostic category embeddings to adapt to the ODOV object detection task.}
	\label{fig:overall}
	\vspace{-15pt}
\end{figure*}

We aim to fully exploit the generalization capability of pretrained VLMs (\eg, CLIP) to handle the ODOV object detection, and propose a baseline method -- O\textbf{D}O\textbf{V} Detec\textbf{tor} (\textbf{DVtor}).
As illustrated in Fig.~\ref{fig:overall}, we adopt a frozen CLIP encoder and propose a novel strategy that dynamically generates customized Domain-category embeddings based on test images, thereby accommodating both category diversity and domain variability.
On the one hand, we introduce the \textit{Domain-Agnostic Category Prompt (DAPmt)} to emphasize intrinsic category semantics while avoiding the limitations imposed by style information.
On the other hand, we design a learnable \textit{Domain Projection and Grafting (DP\&G) network}, which integrates domain-specific features of the input image with DAPmt.

Since the training images originate only from a single source domain, we align prompts with image embeddings during training through implicit domain augmentation and contrastive learning.

\subsection{Domain-Agnostic Category Prompt (DAPmt)}
\label{sec:style}
VLMs such as CLIP associate the visual images with text captions through large-scale pre-training.
CLIP based open-scene detection has achieved high accuracy on multiple datasets, utilizing the manually crafted prompts (\eg, `a photo of \{\}') as the text prompts. 
Such simple prompts are easy to obtain, but can not make full use of the VLMs, which is especially highlighted in the ODOV setting since the same category of object from different domains shows various characteristics.
Some recent works~\cite{gu2021open,zhong2022regionclip,kuo2022f,wu2023clipself,bangalath2022bridging} manually design the prompt templates considering the dataset characteristics. 
For example, using templates like `a black-and-white photo of \{\}' or `a rainy photo of \{\}', which are time-consuming and also difficult to generalize to other datasets.
Although several learnable and automated prompt generation methods~\cite{feng2022promptdet,li2023desco,du2024lami,jin2024llms,kim2024retrieval} have been proposed, they often rely on generic category descriptions, overlooking category-specific attributes and the style information, as their focus remains on conventional OVD tasks.

To address the above challenge of generating low-ambiguity prompts applicable across diverse open domains, we advocate the use of large language models (LLMs) to generate such prompts. Specifically, the LLM produces category-specific yet domain-independent attribute descriptions for each category in the dataset, serving as Domain-Agnostic Category Prompt (DAPmt). This approach avoids the limitations caused by domain shifts and reduces ambiguities arising from category names through rich attribute descriptions. By generating text with fine-grained attributes, it not only enhances the discriminability of similar categories but also preserves generalizability across different domains.

Specifically, for each category, we provide the large language model (LLM, \eg, ChatGPT-4o) with a unified instruction template:
`A/An \{category\} has/is [Appearance feature1], [Appearance feature2], …'.
During generation, we guide the model to ensure that attribute elements are non-redundant, emphasize each category’s distinctive visual properties, and disregard domain information to preserve cross-domain consistency. For example, for the category \{Airplane\}, the generated description is:
`An airplane is a large vehicle with long wings and a streamlined body'.
The resulting descriptions exhibit strong robustness to domain shifts, enabling them to generalize across diverse styles and contexts. At the same time, we deliberately exclude information such as color that is easily influenced by domain factors, reducing interference from illumination or artistic styles. Although the method still requires minor manual adjustments, it significantly reduces the workload compared with traditional approaches that rely heavily on handcrafted prompts. Overall, this strategy aims to produce domain-agnostic category prompts that enable the model to form a general and precise understanding of categories, thereby improving its robustness in ODOV object detection tasks.

\subsection{Domain Projection and Grafting (DP\&G)}
\label{sec:domain}
To further address the ever-changing domain shift problem, we propose \textit{Domain Projection and Grafting (DP\&G)}. The core idea is to extract the \textbf{domain embeddings} (such as style or imaging conditions) of each input image and integrate them with the category text embeddings of DAPmt, producing customized embeddings that jointly capture both domain and category information. In the following, we present the detailed implementation of this strategy.

\textbf{Domain-aware embedding extraction.}
First, we extract the multi-level domain-aware features.
Specifically, during training on single source-domain data, we first apply random perturbations to the features based on the classic AdaIN~\cite{huang2017arbitrary} and NP~\cite{fan2023towards} to simulate domain variations~\footnote{Following the rationale in~\cite{huang2017arbitrary}, the mean and standard deviation in the feature map implicitly represent the domain-aware information.}.
Then, for an input image \textbf{I}, we extract the (disturbed) feature map $\textbf{F}_l \in \mathbb{R}^{W \times H \times C}$ from the \(l^{\text{th}}\) layer of the image encoder $\mathcal{E}$, in which $W, H, C$ denote the height, width, and channel, respectively.
For the \(c^{\text{th}}\) channel, the mean \(\mu_l^c\) and standard deviation \(\sigma_l^c\) are computed as 
$ \scriptstyle
	\mu_l^c = \mathrm{avg} (\textbf{F}^c_{l}({w,h})),
	\sigma_l^c = \sqrt{\mathrm{avg} \left(\textbf{F}^c_{l}({w,h}) - \mu_l^c\right)^2}.
$
We contact all $C$ channels of $\mu_l^c$ and $\sigma_l^c$ to obtain the mean/standard deviation vectors, as $\mathbf{\mu}_l \in \mathbb{R}^{C}$ and $\mathbf{\sigma}_l \in \mathbb{R}^{C}$.
This way, we can obtain the initial domain-aware features by concatenating the mean and standard deviation vectors as
$
\mathcal{E}_l = \left[\mathbf{\mu}_l; \mathbf{\sigma}_l\right].
$

The domain feature of layer $l$ namely \(\mathcal{E}_l\) is then processed through adaptive average pooling and a fully connected layer to align the dimension with that of the DAPmt embeddings (Sec.~\ref{sec:style}). 
As shown in Fig.~\ref{fig:overall}, we extract the domain features from multiple layers. 
These features are then input into an aggregation network to get the final domain embedding $\mathbf{E}_{\text{domain}}$, in which a set of learnable weights dynamically balances the contributions from each layer through weighted integration. 

We then consider to project the domain embeddings $\mathbf{E}_{\text{domain}}$ into a regularized embedding space.
Specifically, we hope the learned domain-aware embeddings are \textit{irrelevant} (orthometric) to the category embeddings.
For this purpose, we apply an orthogonality constraint loss, which minimizes the similarity between the domain embeddings and the category DAPmt embeddings, thereby effectively separating them, as
\begin{equation}
\label{eq:1}
\footnotesize
\mathcal{L}_{\text{orth}} = \sum_{g=1}^{\mathcal{C}^\mathrm{base}} \mathrm{cossim} \left(\mathbf{E}_{\text{domain}}, \mathbf{E}_{\text{category}}^g\right),
\end{equation}
where $\mathbf{E}_{\text{domain}}$ represents the learnable domain embeddings, $\mathbf{E}_{\text{category}}^g$ denotes the DAPmt embeddings for the $g^{\mathrm{th}}$ category, $\mathcal{C}^\mathrm{base}$ is the total number of (base) categories, and $\mathrm{cossim}$ denotes the cosine similarity.

\textbf{Grafting the domain-specific embedding to DAPmt.}
Finally, we develop a \textbf{domain and category embedding fusion (DCF)} module. 
Specifically, we concatenate the domain and category embeddings, followed by a multi-layer perception (MLP) to project it into the fusion space as
\begin{equation}
	\label{eq:2}
    \footnotesize
	\mathbf{E}_{\text{fusion}}^g = \text{MLP}([\mathbf{E}_{\text{domain}}; \mathbf{E}_{\text{category}}^g]),
\end{equation}
which denotes the fused embedding for the \(g\)-th category.

After that, to preserve core semantic information from the category text descriptions, a residual connection of $\mathbf{E}_{\text{category}}^g$ is applied as 
\begin{equation}
\label{eq:3}
\footnotesize
	\mathbf{E}_{\text{DCF}}^g = \alpha \cdot \mathbf{E}_{\text{fusion}}^g + (1 - \alpha) \cdot \mathbf{E}_{\text{category}}^g,
\end{equation}
where a learnable parameter \(\alpha\) within the range \([0,1]\) is used to dynamically balance the direct fusion embedding and the original category embedding $\mathbf{E}_{\text{category}}^g$.

After the DCF module, the proposed method grafts the domain-specific embedding (from the input image) on the domain-agnostic category embedding (from DAPmt), to generate the \textit{\textbf{domain-customized category embeddings for each given image and category prompt}}, thereby improving the ODOV detection ability.

Finally, during training, disturbed visual features extracted from the image encoder form $\mathbf{F}l$, while ROI Align~\cite{he2017mask} generates the object-level features $\mathbf{F}{\mathrm{RoI}}$. The domain–category embedding $\mathbf{E}{\text{DCF}}^g$ is aligned with $\mathbf{F}{\mathrm{RoI}}$ using a contrastive loss as
\begin{equation}
\label{eq:4}
\footnotesize
    \mathcal{L}_{\text{cons}} = 1 - \frac{\mathbf{E}_{\text{DCF}}^g \cdot \mathbf{F}_\mathrm{RoI}}{\|\mathbf{E}_{\text{DCF}}^g\| \|\mathbf{F}_\mathrm{RoI}\|},
\end{equation}
where $\|\cdot\|$ \text{ is the } $L_2$ \text{ norm of a vector.}

\subsection{Implementation Details}
\textbf{Training stage.} 
During training, we apply random perturbations to the mean and standard deviation of the first and second layer features output by the image encoder, and simultaneously, the multi-layer domain features in DP\&G are drawn from layers [3, 5, 7, 11] of ViT-B/16, layers [6, 10, 14, 23] of ViT-L/14, and all layers of ResNet.
Our method is trained using 16 3090 GPUs, with a batch size of 10 per GPU. 
We use AdamW configured with a learning rate of \(10^{-4}\) and a weight decay of 0.1. Training is conducted on the LVIS training set for 50 epochs. 

\textbf{Detailed illustrations and descriptions of testing stage are provided in the supplementary material.}




\section{Experimental Results}
\subsection{Setup}
\textbf{ODOV settings.} 
For the open domain, we train the models on single-domain natural images (LVIS training set) and test on all 15 open-domains in OD-LVIS, similar to the single-domain generalization setting~\cite{qiao2020learning}.
For the open vocabulary, we follow the OV-LVIS in ViLD~\cite{gu2021open} for dataset splitting. Among all categories in OD-LVIS, \ie, in OV-LVIS, 405 `frequent' and 461 `common' categories are assigned as {base categories} for training, while 337 `rare' categories are {novel categories} only for testing.
Note that, OD-LVIS is the \textit{only benchmark meeting the ODOV setting}, so all main experiments are conducted on it.

\textbf{Evaluation methods.} 
To build the benchmark evaluation of the ODOV detection, we selected several mainstream VLM-based OVD methods for comparison on OD-LVIS.
Specifically, we include the transfer learning approaches, \ie, F-VLM~\cite{kuo2022f}, OWL-ViT~\cite{minderer2022simple}, CLIPSelf~\cite{wu2023clipself}, DeCLIP~\cite{wang2025declip}, MM-OVOD~\cite{xu2023exploring}, and OV-DQUO~\cite{wang2025ov}, and several knowledge distillation methods, \ie, RKDWTF~\cite{bangalath2022bridging}, DK-DETR~\cite{li2023distilling}, RegionCLIP~\cite{zhong2022regionclip}, and region-aware training method YOLO-World~\cite{cheng2024yolo}, YOLOE~\cite{wang2025yoloe}.
We also include two recent domain generalization (DG) methods, ALT~\cite{gokhale2023improving}, ABA~\cite{cheng2023adversarial}, NP~\cite{fan2023towards}, MixStyle~\cite{zhou2024mixstyle}, and PhysAug~\cite{xu2025physaug} for comparison. 

\textbf{Evaluation metrics.}
For the evaluation metrics, the average precision on `frequent' and `common' categories, denoted as $AP_f$ and $AP_c$, respectively, serves as the metric for base categories, while the average precision on `rare' categories, denoted as $AP_r$, is used to evaluate novel categories.
The average precision for all categories is denoted as $AP$.

\subsection{Main Results on OD-LVIS}

\begin{table}[t!]
\setlength{\tabcolsep}{3.2pt}
	\centering
    \tiny
	\caption{Comparison with SOTA on OD-LVIS (\%).}\vspace{-10pt}
	\label{tab:OD-LVIS}
    \begin{threeparttable}
	\begin{tabular}{l|c|c|c|c|c|c}
		\toprule
		Method    & Backbone  & Training Data  &$AP_f$     &$AP_c$    &$AP_r$  &$AP$  \\\hline
		\multirow{2}*{RegionCLIP~\cite{zhong2022regionclip}} & {RN50}$^*$  & \multirow{2}*{CC3M}  &  16.6      & 13.0      & 9.7     & 13.9 \\
		  & {RN50x4}$^*$ &  & 19.5       & 15.8      & 12.4     & 16.7 \\
		\hline
		\multirow{2}*{OWL-ViT~\cite{minderer2022simple}} & ViT-B/16  & \multirow{2}*{O365 + VG}  & 13.1       & 13.9      & 13.2     & 13.5\\
		  & ViT-L/14 &   & 22.1      & 21.6      & 19.9     & 21.5  \\
		\hline
		\multirow{4}*{RKDWTF~\cite{bangalath2022bridging}} & {RN50}$^*$ Base  & \multirow{4}*{LVIS-base + IN-L}  & 14.6  & 12.4  & 8.7  & 12.6 \\
		  & {RN50}$^*$ RKDPIS  &  & 13.4 & 12.1 & 10.3 & 12.3 \\
		  & {RN50}$^*$ WTF &  &  14.0  & 12.5 & 11.3 & 12.9 \\
		  & {RN50}$^*$ WTF8x  &  & 15.8 & 14.3 & 11.9 & 14.5 \\
		\hline
        DK-DETR~\cite{li2023distilling} & RN50 & LVIS-all  & 21.1 &  19.4 & 15.3 & 19.4 \\
        \hline 
		  MM-OVOD~\cite{xu2023exploring} & \multirow{4}*{{RN50}$^*$\_Agg}  & \multirow{2}*{\makecell{LVIS-base}} & 20.5 & 19.8  & 14.0 & 19.0 \\
        ~~~ + DAPmt &  &  & 20.8   &20.4   & 14.5  & 19.5  \\
        \cline{1-1}\cline{3-7}
		  MM-OVOD &  &  \multirow{2}*{\makecell{LVIS-base + IN-L}}  & 20.4  & 20.4  & 15.9  & 19.6\\
        ~~~ + DAPmt &  &   & 21.1   &20.9   & 16.0  & 20.1  \\
		\hline
        YOLO-World~\cite{cheng2024yolo} & \multirow{2}*{{YOLOv8-L}$^*$} & \multirow{4}*{O365 + GoldG}  &21.9 &19.1 &19.3 &20.2 \\
        ~~~ + DAPmt &  &  &22.0 &19.6 &19.7 &20.6 \\
        \cline{1-2}\cline{4-7}
        YOLOE~\cite{wang2025yoloe} & \multirow{2}*{{YOLOv11-L}$^*$}   &  &13.7  &8.6  &6.8  &10.3  \\
        ~~~ + DAPmt &   &  &14.3  &9.5  &9.1  &11.3  \\
		\hline
        OV-DQUO~\cite{wang2025ov} & \multirow{2}*{ViT-B/16} & \multirow{24}*{LVIS-base} &12.8  &14.8  &14.8  &14.0  \\
        ~~~ + DAPmt & &  &13.5   &15.3   &15.8   &14.7  \\
        \cline{1-2}\cline{4-7}
        OV-DQUO  & \multirow{2}*{ViT-L/14} & &16.4  &20.6  &21.2  &19.1  \\ 
        ~~~ + DAPmt & & &17.2   &21.5  &22.3   &20.0   \\
		\cline{1-2}\cline{4-7}
		  F-VLM (CLIP)~\cite{kuo2022f}  & \multirow{4}*{RN50x16} & & 16.7 & 14.4 & 13.7 & 15.2  \\
         ~~~ + DAPmt & &  &17.2  &15.3 &14.7 &16.0   \\
         ~~~ + DP\&G & &  &17.5  &16.2 &15.2 &16.5  \\
         \textbf{DVtor (CLIP)} & & &\textbf{17.6}  &\textbf{16.9} &\textbf{15.8} &\textbf{17.0}  \\ 
		\cline{1-2}\cline{4-7}
		F-ViT (CLIPSelf)~\cite{wu2023clipself} & \multirow{4}*{ViT-B/16} & & 17.1 & 12.0  & 12.2 &14.0  \\
        ~~~ + DAPmt & &  &17.5  &12.7 &13.2 &14.7  \\
        ~~~ + DP\&G & &  &18.6  &13.8 &13.7 &15.7  \\
        \textbf{DVtor (CLIPSelf)} &  &  &\textbf{19.0}  &\textbf{14.3} &\textbf{14.0} &\textbf{16.1}  \\ 
        
        \cline{1-2}\cline{4-7}
		  F-ViT (CLIPSelf) & \multirow{4}*{ViT-L/14} & &22.5   & 21.3 & 20.2 & 21.6 \\ 
        ~~~ + DAPmt & &  &22.8  &21.7 &20.8 &22.0  \\ 
		~~~ + DP\&G &  & &23.5  &22.2 &21.5 &22.6  \\ 
        \textbf{DVtor (CLIPSelf)} &  &  &\textbf{23.9}  &\textbf{22.9} &\textbf{21.6} &\textbf{23.1}  \\
        \cline{1-2}\cline{4-7}
        
		F-ViT (DeCLIP)~\cite{wang2025declip} & \multirow{4}*{ViT-B/16}  &  &17.8  &12.9  &13.2  &14.9  \\
        ~~~ + DAPmt  & &   &18.3  &13.6  &14.6  &15.6  \\
        ~~~ + DP\&G  & &   &18.8  &14.1  &14.9  &16.1  \\
        \textbf{DVtor (DeCLIP)}   & &  &\textbf{19.5}  &\textbf{14.7}  &\textbf{15.5}  &\textbf{16.7}  \\
        \cline{1-2}\cline{4-7}

		  F-ViT (DeCLIP) & \multirow{4}*{ViT-L/14} & &23.0  &21.7  &21.4  &22.2  \\
		~~~ + DAPmt &  &  &23.6  &22.0  &22.6  &22.7  \\ 
		~~~ + DP\&G &  &  &24.1  &22.3  &22.8  &23.1  \\ 
        \textbf{DVtor (DeCLIP)} &  &  &\textbf{24.9}  &\textbf{22.6}  &\textbf{23.2}  &\textbf{23.6}  \\ 

		\bottomrule
	\end{tabular} 
        \begin{tablenotes}
            \scriptsize
            \item Notes: IN-L denotes the inclusion of images corresponding to the 997 categories shared between ImageNet-21k-P~\cite{ridnik2021imagenet} and LVIS, `$^*$' indicates that the backbone is not initialized with CLIP, O365 is an abbreviation for Objects365, CC3M, GoldG, and VG are all publicly available datasets.  
        \end{tablenotes}
    \end{threeparttable} \vspace{-17pt}
\end{table}

Table~\ref{tab:OD-LVIS} shows the results of all comparative methods and our DVtor on OD-LVIS.
We can first see that, the proposed DVtor (DeCLIP ViT-L/14) with the DAPmt and DP\&G network, achieves the best performance among all competitors.
Specifically, when using CLIPSelf ViT-L/14 (304.43M) as the backbone, DVtor surpasses F-ViT with the same backbone by 1.4\%, 1.6\%, 1.4\%, and 1.5\% on the frequent, common, rare, and overall categories, respectively.
Moreover, with the DeCLIP ViT-L/14 backbone, DVtor further achieves improvements of 1.9\%, 0.9\%, 1.8\%, and 1.4\% over F-ViT across the same four category groups.
For smaller networks, such as CLIPSelf and DeCLIP with ViT-B/16 (86.26M), our method achieves overall $AP$ improvements of 2.1\% and 1.8\% compared with F-ViT using the same backbone.
In addition, DVtor with RN50×16 (167.33M) surpasses F-VLM with the same backbone by 1.8\% in overall $AP$, and notably, our RN50×16 model (17.0\% @$AP$) even outperforms F-VLM with a much larger RN50×64 backbone (16.9\% @$AP$).
These results demonstrate the remarkable advantages of the DAPmt and DP\&G modules in enhancing generalization for open-domain detection.
Moreover, the overall performance of all methods on OD-LVIS remains relatively low, highlighting the challenging nature of this benchmark and the substantial room for further improvement.


\subsection{Ablation Study}
\textbf{Effectiveness of the proposed DAPmt.} As shown in Table~\ref{tab:OD-LVIS}, when we integrate DAPmt (`+ DAPmt') into F-VLM (CLIP RN50×16), F-ViT (CLIPSelf ViT-B/16 and ViT-L/14), and F-ViT (DeCLIP ViT-B/16 and ViT-L/14), it achieves improvements of 1.0\%, 1.0\%, 0.6\%, 0.6\%, and 1.2\% on the `rare' categories, respectively. 
These results indicate that introducing DAPmt in the training process enables our model to effectively learn the category semantics from CLIP, thus enhancing the model’s generalization capabilities.
Moreover, we apply DAPmt to the inference of various OVD frameworks, including MM-OVOD, YOLO-World, YOLOE, and OV-DQUO, all of which achieve consistent and significant performance improvements, further demonstrating the generality and robustness of DAPmt.

\textbf{Effectiveness of the proposed DP\&G.} 
To validate the DP\&G solely, we train DP\&G based on fixed prompt templates (\eg, `a photo of \{\}') instead of using DAPmt.
As shown in Table~\ref{tab:OD-LVIS}, we integrate the DP\&G network into both F-VLM and F-ViT frameworks (`+ DP\&G'). The results show that, when using CLIPSelf ViT-B/16 and ViT-L/14 as backbones, DVtor surpasses F-ViT by 1.5\%, 1.8\%, 1.5\%, and 1.7\%, as well as by 1.0\%, 0.9\%, 1.3\%, and 1.5\% on the frequent, common, rare, and overall categories, respectively.
Similar performance improvements are also observed when F-ViT adopts DeCLIP ViT-B/16 and ViT-L/14 as backbones, and when F-VLM uses RN50×16 as the backbone.
We attribute these improvements to the DP\&G module’s ability to effectively extract and integrate visual domain information, enriching the semantic representation of customized prompts and further enhancing the model's generalization across diverse scenarios.
Moreover, compared with models using only `DAPmt' or `DP\&G', DVtor (\ie, DAPmt + DP\&G) achieves superior results across all metrics, indicating that removing either module leads to a decline in performance, thereby further verifying the effectiveness and complementarity of both components.

\begin{figure}[t!] 
	\centering 
	\includegraphics[width=0.95\linewidth]{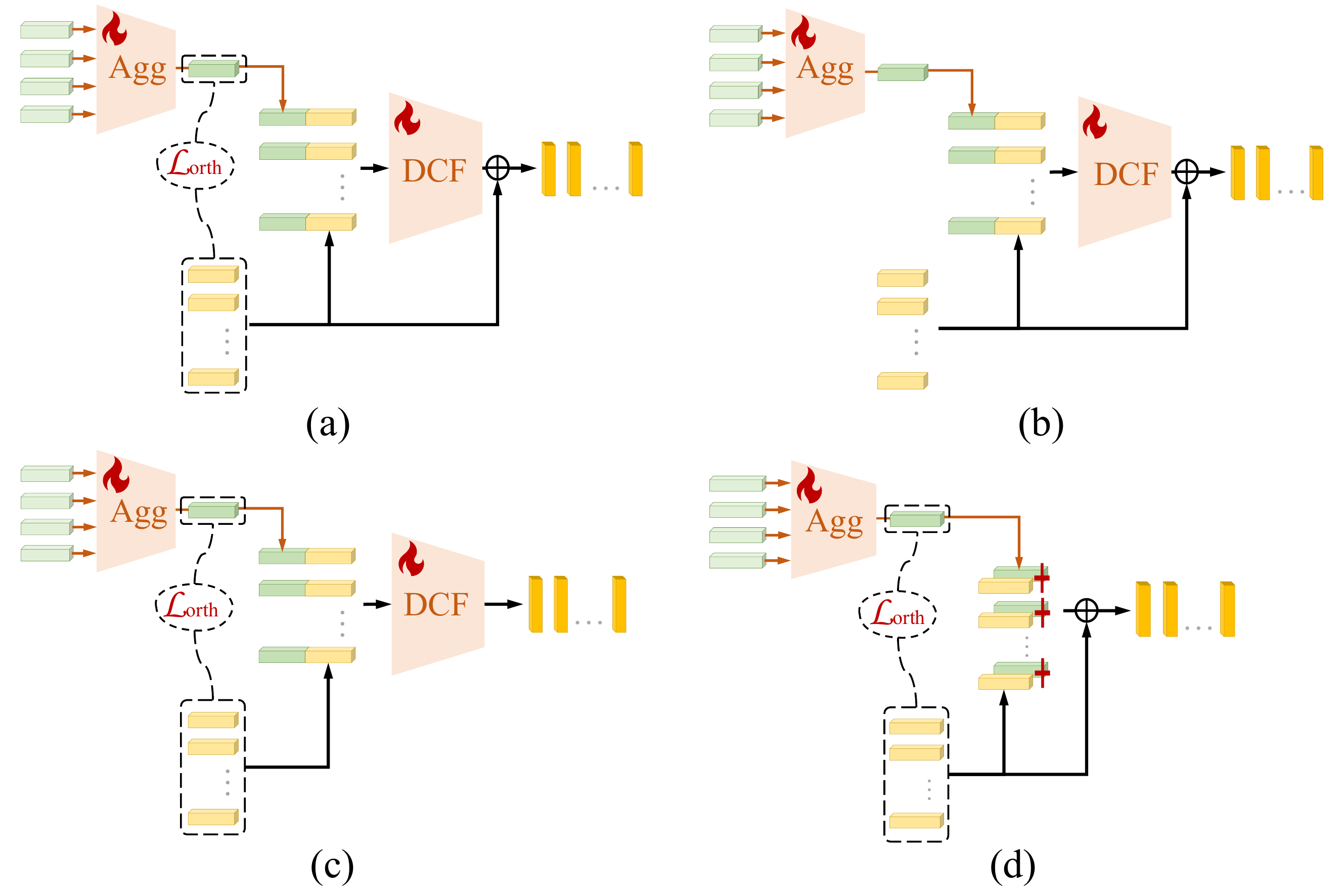}
	\vspace{-10pt}
	\caption{The different fusion structures.}
	\label{fig:stfusion}
    \vspace{-10pt}
\end{figure}

\begin{table}[t!] 
	\centering
	\tiny
	\caption{Comparison of the fusion structures on OD-LVIS (\%).}\vspace{-10pt}
	\label{tab:fusion}
	\begin{tabular}{l|ccccc}
		\toprule
		Method &$\text{Struct.}$  &$AP_f$  &$AP_c$  &$AP_r$  &$AP$ \\
		\hline
    		\multirow{4}*{DVtor (DeCLIP ViT-B/16)}  &$\rm (a)$  &19.5 &14.7 &15.5 &16.7  \\
    		 &$\rm (b)$  &17.6 &12.7 &13.1 &14.7  \\
    		 &$\rm (c)$  &18.1 &13.1 &9.5  &14.4  \\
    		 &$\rm (d)$  &18.0 &11.9 &13.2 &14.5  \\ 
		\bottomrule 
	\end{tabular}  \vspace{-18pt}
\end{table}

\textbf{Ablation study of different structures.} As shown in Fig.~\ref{fig:stfusion}, we compare: {(a)} Ours; {(b)} Removing the orthogonal regularization loss; 
{(c)} Removing the residual connection; {(d)} Replacing the DCF module with a direct weighted summation.
As shown in Table~\ref{tab:fusion}, the structure in (b) results in an overall performance drop. This is because the domain embeddings extracted from the input images may still contain a small amount of object category information, and without maintaining orthogonality, the fusion text features contain reduced category discrimination than the original, which negatively impacts the model's recognition accuracy.
Structure (c) leads to a significant drop in accuracy for novel categories, indicating that the fusion process may cause some semantic information of categories to be lost. 
This indicates that the residual connection helps retain original semantic information, thus enhancing the model's generalization capability for novel categories. 
Structure (d) also causes an accuracy drop. This direct fusion method is too simplistic to effectively integrate the two feature types, resulting in suboptimal performance.



\begin{table}[t!]
\setlength{\tabcolsep}{3.2pt}
	\centering
    \tiny
	\caption{Comparison with OVD methods on OV-LVIS (\%).} \vspace{-10pt}
	\label{tab:LVIS}
    \begin{tabular}{l|c|c|c|c|c|c}
        \toprule
        Method    & Backbone & Training Data  &$AP_f$     &$AP_c$    &$AP_r$  &$AP$  \\\hline
        \multirow{2}*{RegionCLIP~\cite{zhong2022regionclip}} & {RN50}$^*$ &\multirow{2}*{CC3M} & 34.0 & 27.4 & 17.1 &  28.2 \\
          & {RN50x4}$^*$ & & 36.9 & 32.1 & 22.0 & 32.3 \\
        \hline
        \multirow{2}*{OWL-ViT~\cite{minderer2022simple}} & ViT-B/16 & \multirow{2}*{O365 + VG}  & - & - & 20.6 &  27.2 \\
          & ViT-L/14 &  & - & - & 31.2 & 34.6 \\
        \hline
        \multirow{4}*{RKDWTF~\cite{bangalath2022bridging}} & {RN50}$^*$ Base & \multirow{4}*{LVIS-base + IN-L} & 26.4 & 19.4 & 12.2 & 20.9 \\
          & {RN50}$^*$ RKDPIS &  & 25.5 & 20.9 & 17.3 & 22.1 \\
          & {RN50}$^*$ WTF &  & 26.7 & 21.4 & 17.1 & 22.8 \\
          & {RN50}$^*$ WTF8x  &  & 29.1 & 25.0 & 21.1 & 25.9 \\
        \hline
        \multirow{4}*{MM-OVOD~\cite{xu2023exploring}} & {RN50}$^*$\_Avg  & \multirow{2}*{LVIS-base}  & - & - & 20.7 & 30.5 \\
          & {RN50}$^*$\_Agg  &  & - & - & 19.3 & 30.6 \\
          \cline{3-7}
          & {RN50}$^*$\_Avg  & \multirow{2}*{LVIS-base + IN-L} & - & - & 26.5 & 32.8 \\
          & {RN50}$^*$\_Agg  &  & - & - & 27.3 & 33.1 \\
        \hline
        DK-DETR~\cite{li2023distilling} & RN50  & LVIS-all & 40.2 & 32.0 & 22.2 & 33.5 \\
        \hline
        YOLO-World~\cite{cheng2024yolo} & {YOLOv8-L}$^*$ & \multirow{2}*{O365 + GoldG}  &35.4 &24.9 &22.9 &28.7 \\
        \cline{1-2}\cline{4-7}
        YOLOE~\cite{wang2025yoloe} & {YOLOv11-L}$^*$ & &36.5 &35.0 &29.1 &35.2 \\
        \hline 
        \multirow{2}*{OV-DQUO~\cite{wang2025ov}} & ViT-B/16 & \multirow{12}*{\makecell{LVIS-base}} &23.8  &27.7   &29.4  &26.5  \\
          & ViT-L/14 & &28.5  &36.0  &39.5  &33.7  \\ 
        \cline{1-2}\cline{4-7}
        F-VLM (CLIP)~\cite{kuo2022f} & RN50x16 & & - & - & 30.4 & 32.1 \\
        \cline{1-2}\cline{4-7}
          \multirow{2}*{F-ViT (CLIPSelf)}~\cite{wu2023clipself} & ViT-B/16  &  &29.1  &21.8 &25.3 &25.2  \\
          & ViT-L/14 & &35.6  &34.6  &34.9  &35.1  \\
        \cline{1-2}\cline{4-7}
        \multirow{2}*{F-ViT (DeCLIP)~\cite{wang2025declip}} & ViT-B/16 & & 29.8  &22.4  & 26.8 & 26.0 \\
          & ViT-L/14 & & 36.5  & 35.2  & 37.2  &36.0\\ 
        \cline{1-2}\cline{4-7}
        \textbf{DVtor (CLIP)} & RN50x16 & &33.0  & 34.1 & 33.1 & 33.5  \\ 
        \cline{1-2}\cline{4-7}
        \multirow{2}*{\textbf{DVtor (CLIPSelf)}} & ViT-B/16 & &30.4 &23.2 &26.3 &26.6  \\ 
          & ViT-L/14 & &36.9  &35.8  &36.4  &36.3  \\ 
        \cline{1-2}\cline{4-7}
        \multirow{2}*{\textbf{DVtor (DeCLIP)}} & ViT-B/16 & &31.0 &23.7 &28.1 &27.3   \\ 
          & ViT-L/14 & &37.6 &35.9 &39.0 &37.1  \\ 
        \bottomrule    
    \end{tabular}\vspace{-18pt}
\end{table}

\subsection{More Results}
\label{sec:B}
\textbf{Comparison with OVD methods on OV-LVIS~\cite{gu2021open}.} 
We further evaluate our proposed method on the OV-LVIS validation set, as shown in Table~\ref{tab:LVIS}. Our DVtor (based on DeCLIP ViT-L/14) achieves the highest overall $AP$ among all methods. Moreover, when using the same backbone, the proposed method consistently outperforms F-ViT (CLIPSelf and DeCLIP) and F-VLM across all four settings. Overall, the performance improvement of our method on OV-LVIS is relatively modest compared to that on OD-LVIS, mainly because OV-LVIS exhibits lower domain diversity, whereas our method demonstrates stronger competitiveness on the more diverse OD-LVIS benchmark.

\textbf{Comparison with DG Methods on OD-LVIS.} 
In Table~\ref{tab:dg}, we compare our proposed method with five domain generalization (DG) methods based on F-ViT using DeCLIP ViT-B/16 as the backbone on OD-LVIS.
It can be observed that our method achieves the best performance.

\begin{table}[t!]
	\centering
    \tiny
    \caption{Comparison with DG methods (\%).}\vspace{-10pt}
	\label{tab:dg}
	\begin{tabular}{l|cccc}
		\toprule
		Method      &$AP_f$     &$AP_c$    &$AP_r$    &$AP$  \\\hline
        F-ViT (DeCLIP ViT-B/16)                   &17.8  &12.9  &13.2  &14.9 \\
        ~~~ + ALT~\cite{gokhale2023improving}     &18.2  &13.2  &13.1  &15.1 \\
        ~~~ + ABA~\cite{cheng2023adversarial}     &18.0  &13.4  &13.8  &15.3 \\
        ~~~ + NP~\cite{fan2023towards}            &18.6  &14.0  &14.2  &15.8 \\
        ~~~ + MixStyle~\cite{zhou2024mixstyle}    &17.9  &13.3  &13.5  &15.1 \\
        ~~~ + PhysAug~\cite{xu2025physaug}        &18.3  &13.6  &13.7  &15.5 \\
        \textbf{DVtor (DeCLIP)}                   &19.5  &14.7  &15.5  &16.7 \\
		\bottomrule
	\end{tabular} \vspace{-8pt}
\end{table}

\textbf{Cross-Dataset Transfer Results.}
Table~\ref{tab:O365} presents the cross-dataset transfer results from OV-LVIS to Objects365~\cite{shao2019objects365}. We compare our proposed method with Detic~\cite{zhou2022detecting}, MM-OVOD~\cite{kaul2023multi}, F-VLM~\cite{kuo2022f}, and F-ViT (CLIPSelf~\cite{wu2023clipself} and DeCLIP~\cite{wang2025declip}), using the standard bounding box $AP$ metric on Objects365 for evaluation.
In all experiments, Detic and MM-OVOD are trained on LVIS-all, where IN-L models use ImageNet-21k-P as additional weak supervision, while F-ViT (CLIPSelf and DeCLIP) and our method are trained on the base categories of OV-LVIS and evaluated on the Objects365 validation set. Following MM-OVOD, we define the bottom one-third of categories in Objects365, ranked by frequency, as rare categories.
When using CLIP RN50×16 as the backbone, our method improves over F-VLM by 1.3\% in $AP_r$ and 2.3\% in $AP^\textit{50}$, surpasses F-ViT (CLIPSelf) by 0.7\% and 0.5\%, and outperforms F-ViT (DeCLIP) by 1.1\% and 2.3\%, respectively. With ViT-L/14 as the backbone, our method also achieves notable improvements over both F-ViT (CLIPSelf) and F-ViT (DeCLIP).
Overall, these results demonstrate that our method exhibits significant advantages and strong generalization capabilities in cross-dataset transfer tasks.

\begin{table}[t!]
	\centering
    \tiny
	\caption{Cross-dataset main results on Objects365 (\%).} \vspace{-10pt}
	\label{tab:O365}
	\begin{tabular}{l|c|c|c|c|c}
        \toprule
        Method  &Backbone     & Training Data   &$AP_r$    &$AP$   &$AP^\textit{50}$\\\hline
        \multirow{2}*{Detic~\cite{zhou2022detecting}}   & \multirow{4}*{{RN50}$^*$}  & LVIS-all    & 9.5   & 13.9  &19.7  \\
        &  & LVIS-all + IN-L   &12.4    & 15.6  &22.2  \\
        \cline{1-1}\cline{3-6}
        \multirow{2}*{MM-OVOD~\cite{xu2023exploring}} &   & LVIS-all   & 10.1 & 14.8  &21.0    \\
         &  & LVIS-all + IN-L   & 13.1 & 16.6  & 23.1   \\
        \hline
        F-VLM (CLIP)~\cite{kuo2022f} & RN50x16   &\multirow{10}*{LVIS-base}  & 14.9   &16.2  &25.3    \\
        \cline{1-2}\cline{4-6}
	    \multirow{2}*{F-ViT (CLIPSelf)~\cite{wu2023clipself}} & ViT-B/16    &    &16.8  & 19.0  &32.3    \\
	    & ViT-L/14   &   &21.7  &23.7  &39.2  \\
        \cline{1-2}\cline{4-6}
        \multirow{2}*{F-ViT (DeCLIP)~\cite{wang2025declip}} & ViT-B/16  &   & 17.6  & 20.2   &  33.1  \\
	    & ViT-L/14   &   & 22.3  & 24.5  & 39.8  \\
        \cline{1-2}\cline{4-6}
        \textbf{DVtor (CLIP)} & RN50x16   &  &16.2  &17.9  &27.6    \\\cline{1-2}\cline{4-6}
        \multirow{2}*{\textbf{DVtor (CLIPSelf)}} & ViT-B/16   &  &17.5  &19.2  &32.8    \\
         & ViT-L/14   &  &22.3  &24.0  &39.7    \\\cline{1-2}\cline{4-6}
        \multirow{2}*{\textbf{DVtor (DeCLIP)}} & ViT-B/16   &  &18.7 &21.1  &35.4    \\
         & ViT-L/14   &  &23.7  &25.0  &40.9    \\
	\bottomrule    
	\end{tabular} 
    \vspace{-18pt}
\end{table}

\textbf{Please refer to the supplementary material for visualized results and analysis, as well as the limitations.}

\section{Conclusion}
We have proposed to study a new yet practical problem of ODOV object detection, by considering both the domain and category shifts.
For this purpose, we construct the benchmark OD-LVIS containing 15 domains and 1,203 categories.
We have also developed a baseline method for ODOV detection, which can generate the domain-agnostic text prompts for category embedding, as well as the domain embeddings using a domain projection and grafting network.
By combining both of them, we obtain the customized domain-specific category embedding for each test image, which well adapts ODOV detection.
We provide the benchmark evaluation of a series of SOTA methods on OD-LVIS.
Through these efforts, we hope to pave the way for the study of this new yet significant problem.

\section*{Acknowledgements}
This work was supported in part by the National Natural Science Foundation of China under Grants U2574216 and 62402490; in part by the Emerging Frontiers Cultivation Program of Tianjin University Interdisciplinary Center; in part by the Guangdong Basic and Applied Basic Research Foundation under Grant 2025A1515010101.

{
    \small
    \bibliographystyle{ieeenat_fullname}
    \bibliography{main}

@String(ICLR = {Int. Conf. Learn. Represent.})

@String(AAAI = {AAAI})

@String(ICLR  = {ICLR})

@article{shu2022test,
	title={Test-time prompt tuning for zero-shot generalization in vision-language models},
	author={Shu, Manli and Nie, Weili and Huang, De-An and Yu, Zhiding and Goldstein, Tom and Anandkumar, Anima and Xiao, Chaowei},
	journal={Advances in Neural Information Processing Systems},
	volume={35},
	pages={14274--14289},
	year={2022}
}

@article{zhou2022learning,
	title={Learning to prompt for vision-language models},
	author={Zhou, Kaiyang and Yang, Jingkang and Loy, Chen Change and Liu, Ziwei},
	journal={International Journal of Computer Vision},
	volume={130},
	number={9},
	pages={2337--2348},
	year={2022},
}

@inproceedings{zhou2022conditional,
	title={Conditional prompt learning for vision-language models},
	author={Zhou, Kaiyang and Yang, Jingkang and Loy, Chen Change and Liu, Ziwei},
	booktitle={Proceedings of the IEEE/CVF conference on computer vision and pattern recognition},
	pages={16816--16825},
	year={2022}
}

@inproceedings{khattak2023maple,
	title={Maple: Multi-modal prompt learning},
	author={Khattak, Muhammad Uzair and Rasheed, Hanoona and Maaz, Muhammad and Khan, Salman and Khan, Fahad Shahbaz},
	booktitle={Proceedings of the IEEE/CVF Conference on Computer Vision and Pattern Recognition},
	pages={19113--19122},
	year={2023}
}

@inproceedings{feng2022promptdet,
  title={Promptdet: Towards open-vocabulary detection using uncurated images},
  author={Feng, Chengjian and Zhong, Yujie and Jie, Zequn and Chu, Xiangxiang and Ren, Haibing and Wei, Xiaolin and Xie, Weidi and Ma, Lin},
  booktitle={European conference on computer vision},
  pages={701--717},
  year={2022},
  organization={Springer}
}

@article{li2023desco,
  title={Desco: Learning object recognition with rich language descriptions},
  author={Li, Liunian and Dou, Zi-Yi and Peng, Nanyun and Chang, Kai-Wei},
  journal={Advances in Neural Information Processing Systems},
  volume={36},
  pages={37511--37526},
  year={2023}
}

@inproceedings{du2024lami,
  title={LaMI-DETR: Open-Vocabulary Detection with Language Model Instruction},
  author={Du, Penghui and Wang, Yu and Sun, Yifan and Wang, Luting and Liao, Yue and Zhang, Gang and Ding, Errui and Wang, Yan and Wang, Jingdong and Liu, Si},
  booktitle={European Conference on Computer Vision},
  pages={312--328},
  year={2024},
  organization={Springer}
}

@article{jin2024llms,
  title={Llms meet vlms: Boost open vocabulary object detection with fine-grained descriptors},
  author={Jin, Sheng and Jiang, Xueying and Huang, Jiaxing and Lu, Lewei and Lu, Shijian},
  journal={arXiv preprint arXiv:2402.04630},
  year={2024}
}

@inproceedings{kim2024retrieval,
  title={Retrieval-augmented open-vocabulary object detection},
  author={Kim, Jooyeon and Cho, Eulrang and Kim, Sehyung and Kim, Hyunwoo J},
  booktitle={Proceedings of the IEEE/CVF Conference on Computer Vision and Pattern Recognition},
  pages={17427--17436},
  year={2024}
}

@inproceedings{zareian2021open,
	title={Open-vocabulary object detection using captions},
	author={Zareian, Alireza and Rosa, Kevin Dela and Hu, Derek Hao and Chang, Shih-Fu},
	booktitle={Proceedings of the IEEE/CVF Conference on Computer Vision and Pattern Recognition},
	pages={14393--14402},
	year={2021}
}

@article{kuo2022f,
	title={F-vlm: Open-vocabulary object detection upon frozen vision and language models},
	author={Kuo, Weicheng and Cui, Yin and Gu, Xiuye and Piergiovanni, AJ and Angelova, Anelia},
	journal={arXiv preprint arXiv:2209.15639},
	year={2022}
}

@article{minderer2022simple,
	title={Simple open-vocabulary object detection with vision transformers. arxiv 2022},
	author={Minderer, M and Gritsenko, A and Stone, A and Neumann, M and Weissenborn, D and Dosovitskiy, A and Mahendran, A and Arnab, A and Dehghani, M and Shen, Z and others},
	journal={arXiv preprint arXiv:2205.06230},
	volume={2},
	year={2022}
}

@inproceedings{kim2023contrastive,
	title={Contrastive feature masking open-vocabulary vision transformer},
	author={Kim, Dahun and Angelova, Anelia and Kuo, Weicheng},
	booktitle={Proceedings of the IEEE/CVF International Conference on Computer Vision},
	pages={15602--15612},
	year={2023}
}

@article{kim2023detection,
	title={Detection-oriented image-text pretraining for open-vocabulary detection},
	author={Kim, Dahun and Angelova, Anelia and Kuo, Weicheng},
	journal={arXiv preprint arXiv:2310.00161},
	year={2023}
}

@inproceedings{kim2023region,
	title={Region-aware pretraining for open-vocabulary object detection with vision transformers},
	author={Kim, Dahun and Angelova, Anelia and Kuo, Weicheng},
	booktitle={Proceedings of the IEEE/CVF conference on computer vision and pattern recognition},
	pages={11144--11154},
	year={2023}
}

@inproceedings{wu2023cora,
	title={Cora: Adapting clip for open-vocabulary detection with region prompting and anchor pre-matching},
	author={Wu, Xiaoshi and Zhu, Feng and Zhao, Rui and Li, Hongsheng},
	booktitle={Proceedings of the IEEE/CVF conference on computer vision and pattern recognition},
	pages={7031--7040},
	year={2023}
}

@article{song2023prompt,
	title={Prompt-guided transformers for end-to-end open-vocabulary object detection},
	author={Song, Hwanjun and Bang, Jihwan},
	journal={arXiv preprint arXiv:2303.14386},
	year={2023}
}

@inproceedings{zhong2022regionclip,
	title={Regionclip: Region-based language-image pretraining},
	author={Zhong, Yiwu and Yang, Jianwei and Zhang, Pengchuan and Li, Chunyuan and Codella, Noel and Li, Liunian Harold and Zhou, Luowei and Dai, Xiyang and Yuan, Lu and Li, Yin and others},
	booktitle={Proceedings of the IEEE/CVF conference on computer vision and pattern recognition},
	pages={16793--16803},
	year={2022}
}

@inproceedings{jeong2024proxydet,
	title={ProxyDet: Synthesizing Proxy Novel Classes via Classwise Mixup for Open-Vocabulary Object Detection},
	author={Jeong, Joonhyun and Park, Geondo and Yoo, Jayeon and Jung, Hyungsik and Kim, Heesu},
	booktitle={Proceedings of the AAAI Conference on Artificial Intelligence},
	volume={38},
	number={3},
	pages={2462--2470},
	year={2024}
}

@article{ma2024codet,
	title={Codet: Co-occurrence guided region-word alignment for open-vocabulary object detection},
	author={Ma, Chuofan and Jiang, Yi and Wen, Xin and Yuan, Zehuan and Qi, Xiaojuan},
	journal={Advances in neural information processing systems},
	volume={36},
	year={2024}
}

@inproceedings{kaul2023multi,
	title={Multi-modal classifiers for open-vocabulary object detection},
	author={Kaul, Prannay and Xie, Weidi and Zisserman, Andrew},
	booktitle={International Conference on Machine Learning},
	pages={15946--15969},
	year={2023},
}

@inproceedings{zhou2022detecting,
	title={Detecting twenty-thousand classes using image-level supervision},
	author={Zhou, Xingyi and Girdhar, Rohit and Joulin, Armand and Kr{\"a}henb{\"u}hl, Philipp and Misra, Ishan},
	booktitle={European Conference on Computer Vision},
	pages={350--368},
	year={2022},
}

@article{gu2021open,
	title={Open-vocabulary object detection via vision and language knowledge distillation},
	author={Gu, Xiuye and Lin, Tsung-Yi and Kuo, Weicheng and Cui, Yin},
	journal={arXiv preprint arXiv:2104.13921},
	year={2021}
}

@inproceedings{pham2024lp,
	title={Lp-ovod: Open-vocabulary object detection by linear probing},
	author={Pham, Chau and Vu, Truong and Nguyen, Khoi},
	booktitle={Proceedings of the IEEE/CVF Winter Conference on Applications of Computer Vision},
	pages={779--788},
	year={2024}
}

@article{bangalath2022bridging,
	title={Bridging the gap between object and image-level representations for open-vocabulary detection},
	author={Bangalath, Hanoona and Maaz, Muhammad and Khattak, Muhammad Uzair and Khan, Salman H and Shahbaz Khan, Fahad},
	journal={Advances in Neural Information Processing Systems},
	volume={35},
	pages={33781--33794},
	year={2022}
}

@inproceedings{li2023distilling,
	title={Distilling detr with visual-linguistic knowledge for open-vocabulary object detection},
	author={Li, Liangqi and Miao, Jiaxu and Shi, Dahu and Tan, Wenming and Ren, Ye and Yang, Yi and Pu, Shiliang},
	booktitle={Proceedings of the IEEE/CVF International Conference on Computer Vision},
	pages={6501--6510},
	year={2023}
}

@inproceedings{du2022learning,
	title={Learning to prompt for open-vocabulary object detection with vision-language model},
	author={Du, Yu and Wei, Fangyun and Zhang, Zihe and Shi, Miaojing and Gao, Yue and Li, Guoqi},
	booktitle={Proceedings of the IEEE/CVF Conference on Computer Vision and Pattern Recognition},
	pages={14084--14093},
	year={2022}
}

@article{wu2023clipself,
	title={Clipself: Vision transformer distills itself for open-vocabulary dense prediction},
	author={Wu, Size and Zhang, Wenwei and Xu, Lumin and Jin, Sheng and Li, Xiangtai and Liu, Wentao and Loy, Chen Change},
	journal={arXiv preprint arXiv:2310.01403},
	year={2023}
}

@inproceedings{lin2021domain,
	title={Domain-invariant disentangled network for generalizable object detection},
	author={Lin, Chuang and Yuan, Zehuan and Zhao, Sicheng and Sun, Peize and Wang, Changhu and Cai, Jianfei},
	booktitle={Proceedings of the IEEE/CVF international conference on computer vision},
	pages={8771--8780},
	year={2021}
}

@article{zhang2022gated,
	title={Gated domain-invariant feature disentanglement for domain generalizable object detection},
	author={Zhang, Haozhuo and Yu, Huimin and Yan, Yuming and Wang, Runfa},
	journal={arXiv preprint arXiv:2203.11432},
	year={2022}
}

@inproceedings{wu2022single,
	title={Single-domain generalized object detection in urban scene via cyclic-disentangled self-distillation},
	author={Wu, Aming and Deng, Cheng},
	booktitle={Proceedings of the IEEE/CVF Conference on computer vision and pattern recognition},
	pages={847--856},
	year={2022}
}

@inproceedings{vidit2023clip,
	title={Clip the gap: A single domain generalization approach for object detection},
	author={Vidit, Vidit and Engilberge, Martin and Salzmann, Mathieu},
	booktitle={Proceedings of the IEEE/CVF conference on computer vision and pattern recognition},
	pages={3219--3229},
	year={2023}
}

@article{rao2023srcd,
	title={Srcd: Semantic reasoning with compound domains for single-domain generalized object detection},
	author={Rao, Zhijie and Guo, Jingcai and Tang, Luyao and Huang, Yue and Ding, Xinghao and Guo, Song},
	journal={arXiv preprint arXiv:2307.01750},
	year={2023}
}

@inproceedings{wu2024g,
	title={G-NAS: Generalizable Neural Architecture Search for Single Domain Generalization Object Detection},
	author={Wu, Fan and Gao, Jinling and Hong, Lanqing and Wang, Xinbing and Zhou, Chenghu and Ye, Nanyang},
	booktitle={Proceedings of the AAAI Conference on Artificial Intelligence},
	volume={38},
	number={6},
	pages={5958--5966},
	year={2024}
}

@inproceedings{zhai2022lit,
	title={Lit: Zero-shot transfer with locked-image text tuning},
	author={Zhai, Xiaohua and Wang, Xiao and Mustafa, Basil and Steiner, Andreas and Keysers, Daniel and Kolesnikov, Alexander and Beyer, Lucas},
	booktitle={Proceedings of the IEEE/CVF conference on computer vision and pattern recognition},
	pages={18123--18133},
	year={2022}
}

@article{yao2021filip,
	title={Filip: Fine-grained interactive language-image pre-training},
	author={Yao, Lewei and Huang, Runhui and Hou, Lu and Lu, Guansong and Niu, Minzhe and Xu, Hang and Liang, Xiaodan and Li, Zhenguo and Jiang, Xin and Xu, Chunjing},
	journal={arXiv preprint arXiv:2111.07783},
	year={2021}
}

@article{yuan2021florence,
	title={Florence: A new foundation model for computer vision},
	author={Yuan, Lu and Chen, Dongdong and Chen, Yi-Ling and Codella, Noel and Dai, Xiyang and Gao, Jianfeng and Hu, Houdong and Huang, Xuedong and Li, Boxin and Li, Chunyuan and others},
	journal={arXiv preprint arXiv:2111.11432},
	year={2021}
}

@inproceedings{radford2021learning,
	title={Learning transferable visual models from natural language supervision},
	author={Radford, Alec and Kim, Jong Wook and Hallacy, Chris and Ramesh, Aditya and Goh, Gabriel and Agarwal, Sandhini and Sastry, Girish and Askell, Amanda and Mishkin, Pamela and Clark, Jack and others},
	booktitle={International conference on machine learning},
	pages={8748--8763},
	year={2021},
}

@inproceedings{jia2021scaling,
	title={Scaling up visual and vision-language representation learning with noisy text supervision},
	author={Jia, Chao and Yang, Yinfei and Xia, Ye and Chen, Yi-Ting and Parekh, Zarana and Pham, Hieu and Le, Quoc and Sung, Yun-Hsuan and Li, Zhen and Duerig, Tom},
	booktitle={International conference on machine learning},
	pages={4904--4916},
	year={2021},
}

@article{sun2023eva,
	title={Eva-clip: Improved training techniques for clip at scale},
	author={Sun, Quan and Fang, Yuxin and Wu, Ledell and Wang, Xinlong and Cao, Yue},
	journal={arXiv preprint arXiv:2303.15389},
	year={2023}
}

@article{li2022elevater,
  title={Elevater: A benchmark and toolkit for evaluating language-augmented visual models},
  author={Li, Chunyuan and Liu, Haotian and Li, Liunian and Zhang, Pengchuan and Aneja, Jyoti and Yang, Jianwei and Jin, Ping and Hu, Houdong and Liu, Zicheng and Lee, Yong Jae and others},
  journal={Advances in Neural Information Processing Systems},
  volume={35},
  pages={9287--9301},
  year={2022}
}

@inproceedings{inoue2018cross,
	title={Cross-domain weakly-supervised object detection through progressive domain adaptation},
	author={Inoue, Naoto and Furuta, Ryosuke and Yamasaki, Toshihiko and Aizawa, Kiyoharu},
	booktitle={Proceedings of the IEEE/CVF conference on computer vision and pattern recognition},
	pages={5001--5009},
	year={2018}
}

@inproceedings{cordts2016cityscapes,
	title={The cityscapes dataset for semantic urban scene understanding},
	author={Cordts, Marius and Omran, Mohamed and Ramos, Sebastian and Rehfeld, Timo and Enzweiler, Markus and Benenson, Rodrigo and Franke, Uwe and Roth, Stefan and Schiele, Bernt},
	booktitle={Proceedings of the IEEE/CVF conference on computer vision and pattern recognition},
	pages={3213--3223},
	year={2016}
}

@inproceedings{yu2020bdd100k,
	title={Bdd100k: A diverse driving dataset for heterogeneous multitask learning},
	author={Yu, Fisher and Chen, Haofeng and Wang, Xin and Xian, Wenqi and Chen, Yingying and Liu, Fangchen and Madhavan, Vashisht and Darrell, Trevor},
	booktitle={Proceedings of the IEEE/CVF conference on computer vision and pattern recognition},
	pages={2636--2645},
	year={2020}
}

@article{sakaridis2018semantic,
	title={Semantic foggy scene understanding with synthetic data},
	author={Sakaridis, Christos and Dai, Dengxin and Van Gool, Luc},
	journal={International Journal of Computer Vision},
	volume={126},
	pages={973--992},
	year={2018}
}

@inproceedings{nada2018pushing,
	title={Pushing the limits of unconstrained face detection: a challenge dataset and baseline results},
	author={Nada, Hajime and Sindagi, Vishwanath A and Zhang, He and Patel, Vishal M},
	booktitle={2018 IEEE 9th International Conference on Biometrics Theory, Applications and Systems (BTAS)},
	pages={1--10},
	year={2018}
}

@article{li2018benchmarking,
	title={Benchmarking single-image dehazing and beyond},
	author={Li, Boyi and Ren, Wenqi and Fu, Dengpan and Tao, Dacheng and Feng, Dan and Zeng, Wenjun and Wang, Zhangyang},
	journal={IEEE Transactions on Image Processing},
	volume={28},
	pages={492--505},
	year={2018}
}

@article{johnson2016driving,
	title={Driving in the matrix: Can virtual worlds replace human-generated annotations for real world tasks?},
	author={Johnson-Roberson, Matthew and Barto, Charles and Mehta, Rounak and Sridhar, Sharath Nittur and Rosaen, Karl and Vasudevan, Ram},
	journal={arXiv preprint arXiv:1610.01983},
	year={2016}
}

@inproceedings{yang2016wider,
	title={Wider face: A face detection benchmark},
	author={Yang, Shuo and Luo, Ping and Loy, Chen-Change and Tang, Xiaoou},
	booktitle={Proceedings of the IEEE/CVF conference on computer vision and pattern recognition},
	pages={5525--5533},
	year={2016}
}

@inproceedings{lin2014microsoft,
	title={Microsoft coco: Common objects in context},
	author={Lin, Tsung-Yi and Maire, Michael and Belongie, Serge and Hays, James and Perona, Pietro and Ramanan, Deva and Doll{\'a}r, Piotr and Zitnick, C Lawrence},
	booktitle={Computer Vision--ECCV 2014: 13th European Conference, Zurich, Switzerland, September 6-12, 2014, Proceedings, Part V 13},
	pages={740--755},
	year={2014}
}

@inproceedings{gupta2019lvis,
	title={Lvis: A dataset for large vocabulary instance segmentation},
	author={Gupta, Agrim and Dollar, Piotr and Girshick, Ross},
	booktitle={Proceedings of the IEEE/CVF conference on computer vision and pattern recognition},
	pages={5356--5364},
	year={2019}
}

@inproceedings{zhang2024rethinking,
	title={Rethinking the One-shot Object Detection: Cross-Domain Object Search},
	author={Zhang, Yupeng and Zheng, Shuqi and Han, Ruize and Feng, Yuzhong and Hou, Junhui and Song, Linqi and Feng, Wei and Wan, Liang},
	booktitle={ACM Multimedia 2024}
}

@inproceedings{huang2017arbitrary,
	title={Arbitrary style transfer in real-time with adaptive instance normalization},
	author={Huang, Xun and Belongie, Serge},
	booktitle={Proceedings of the IEEE international conference on computer vision},
	pages={1501--1510},
	year={2017}
}

@article{xu2023exploring,
	title={Exploring multi-modal contextual knowledge for open-vocabulary object detection},
	author={Xu, Yifan and Zhang, Mengdan and Yang, Xiaoshan and Xu, Changsheng},
	journal={arXiv preprint arXiv:2308.15846},
	year={2023}
}

@inproceedings{fan2023towards,
	title={Towards robust object detection invariant to real-world domain shifts},
	author={Fan, Qi and Segu, Mattia and Tai, Yu-Wing and Yu, Fisher and Tang, Chi-Keung and Schiele, Bernt and Dai, Dengxin},
	booktitle={The Eleventh International Conference on Learning Representations (ICLR 2023)},
	year={2023},
}

@inproceedings{qiao2020learning,
	title={Learning to learn single domain generalization},
	author={Qiao, Fengchun and Zhao, Long and Peng, Xi},
	booktitle={Proceedings of the IEEE/CVF conference on computer vision and pattern recognition},
	pages={12556--12565},
	year={2020}
}

@article{hsieh2019one,
	title={One-shot object detection with co-attention and co-excitation},
	author={Hsieh, Ting-I and Lo, Yi-Chen and Chen, Hwann-Tzong and Liu, Tyng-Luh},
	journal={Advances in neural information processing systems},
	volume={32},
	pages={1--10},
	year={2019}
}

@inproceedings{chen2021adaptive,
	title={Adaptive image transformer for one-shot object detection},
	author={Chen, Ding-Jie and Hsieh, He-Yen and Liu, Tyng-Luh},
	booktitle={Proceedings of the IEEE/CVF Conference on Computer Vision and Pattern Recognition},
	pages={12247--12256},
	year={2021}
}

@inproceedings{zhao2022semantic,
	title={Semantic-aligned fusion transformer for one-shot object detection},
	author={Zhao, Yizhou and Guo, Xun and Lu, Yan},
	booktitle={Proceedings of the IEEE/CVF Conference on Computer Vision and Pattern Recognition},
	pages={7601--7611},
	year={2022}
}

@inproceedings{yang2022balanced,
	title={Balanced and hierarchical relation learning for one-shot object detection},
	author={Yang, Hanqing and Cai, Sijia and Sheng, Hualian and Deng, Bing and Huang, Jianqiang and Hua, Xian-Sheng and Tang, Yong and Zhang, Yu},
	booktitle={Proceedings of the IEEE/CVF Conference on Computer Vision and Pattern Recognition},
	pages={7591--7600},
	year={2022}
}

@inproceedings{he2017mask,
	title={Mask r-cnn},
	author={He, Kaiming and Gkioxari, Georgia and Doll{\'a}r, Piotr and Girshick, Ross},
	booktitle={Proceedings of the IEEE international conference on computer vision},
	pages={2961--2969},
	year={2017}
}

@inproceedings{wang2023detecting,
	title={Detecting everything in the open world: Towards universal object detection},
	author={Wang, Zhenyu and Li, Yali and Chen, Xi and Lim, Ser-Nam and Torralba, Antonio and Zhao, Hengshuang and Wang, Shengjin},
	booktitle={Proceedings of the IEEE/CVF Conference on Computer Vision and Pattern Recognition},
	pages={11433--11443},
	year={2023}
}

@inproceedings{ma2022rethinking,
	title={Rethinking open-world object detection in autonomous driving scenarios},
	author={Ma, Zeyu and Yang, Yang and Wang, Guoqing and Xu, Xing and Shen, Heng Tao and Zhang, Mingxing},
	booktitle={Proceedings of the 30th ACM International Conference on Multimedia},
	pages={1279--1288},
	year={2022}
}

@inproceedings{mullappilly2024semi,
	title={Semi-supervised Open-World Object Detection},
	author={Mullappilly, Sahal Shaji and Gehlot, Abhishek Singh and Anwer, Rao Muhammad and Khan, Fahad Shahbaz and Cholakkal, Hisham},
	booktitle={Proceedings of the AAAI Conference on Artificial Intelligence},
	volume={38},
	number={5},
	pages={4305--4314},
	year={2024}
}

@inproceedings{wang2022togethernet,
	title={Togethernet: Bridging image restoration and object detection together via dynamic enhancement learning},
	author={Wang, Yongzhen and Yan, Xuefeng and Zhang, Kaiwen and Gong, Lina and Xie, Haoran and Wang, Fu Lee and Wei, Mingqiang},
	booktitle={Computer Graphics Forum},
	volume={41},
	number={7},
	pages={465--476},
	year={2022},
}

@article{wang2024joint,
	title={Joint image restoration for object detection in snowy weather},
	author={Wang, Jing and Xu, Meimei and Xue, Huazhu and Huo, Zhanqiang and Luo, Fen},
	journal={IET Computer Vision},
	year={2024},
}

@article{wang2024degradation,
	title={Degradation Modeling for Restoration-enhanced Object Detection in Adverse Weather Scenes},
	author={Wang, Xiaofeng and Liu, Xiao and Yang, Hong and Wang, Zhengyong and Wen, Xiaoyue and He, Xiaohai and Qing, Linbo and Chen, Honggang},
	journal={IEEE Transactions on Intelligent Vehicles},
	year={2024},
}

@inproceedings{shu2023clipood,
	title={Clipood: Generalizing clip to out-of-distributions},
	author={Shu, Yang and Guo, Xingzhuo and Wu, Jialong and Wang, Ximei and Wang, Jianmin and Long, Mingsheng},
	booktitle={International Conference on Machine Learning},
	pages={31716--31731},
	year={2023},
}

@inproceedings{dollar2009pedestrian,
	title={Pedestrian detection: A benchmark},
	author={Doll{\'a}r, Piotr and Wojek, Christian and Schiele, Bernt and Perona, Pietro},
	booktitle={2009 IEEE conference on computer vision and pattern recognition},
	pages={304--311},
	year={2009},
}

@inproceedings{zhang2017citypersons,
	title={Citypersons: A diverse dataset for pedestrian detection},
	author={Zhang, Shanshan and Benenson, Rodrigo and Schiele, Bernt},
	booktitle={Proceedings of the IEEE conference on computer vision and pattern recognition},
	pages={3213--3221},
	year={2017}
}

@ARTICLE{9247499,
	author={Pang, Yanwei and Cao, Jiale and Li, Yazhao and Xie, Jin and Sun, Hanqing and Gong, Jinfeng},
	journal={IEEE Transactions on Image Processing}, 
	title={TJU-DHD: A Diverse High-Resolution Dataset for Object Detection}, 
	year={2021},
	volume={30},
	pages={207-219},
}

@inproceedings{cheng2024yolo,
  title={Yolo-world: Real-time open-vocabulary object detection},
  author={Cheng, Tianheng and Song, Lin and Ge, Yixiao and Liu, Wenyu and Wang, Xinggang and Shan, Ying},
  booktitle={Proceedings of the IEEE/CVF Conference on Computer Vision and Pattern Recognition},
  pages={16901--16911},
  year={2024}
}

@inproceedings{gokhale2023improving,
	title={Improving diversity with adversarially learned transformations for domain generalization},
	author={Gokhale, Tejas and Anirudh, Rushil and Thiagarajan, Jayaraman J and Kailkhura, Bhavya and Baral, Chitta and Yang, Yezhou},
	booktitle={Proceedings of the IEEE/CVF Winter Conference on Applications of Computer Vision},
	pages={434--443},
	year={2023}
}

@inproceedings{cheng2023adversarial,
	title={Adversarial Bayesian Augmentation for Single-Source Domain Generalization},
	author={Cheng, Sheng and Gokhale, Tejas and Yang, Yezhou},
	booktitle={Proceedings of the IEEE/CVF International Conference on Computer Vision},
	pages={11400--11410},
	year={2023}
}

@inproceedings{wang2025ov,
  title={Ov-dquo: Open-vocabulary detr with denoising text query training and open-world unknown objects supervision},
  author={Wang, Junjie and Chen, Bin and Kang, Bin and Li, Yulin and Xian, Weizhi and Chen, Yichi and Xu, Yong},
  booktitle={Proceedings of the AAAI Conference on Artificial Intelligence},
  pages={7762--7770},
  year={2025}
}

@inproceedings{wang2025declip,
  title={Declip: Decoupled learning for open-vocabulary dense perception},
  author={Wang, Junjie and Chen, Bin and Li, Yulin and Kang, Bin and Chen, Yichi and Tian, Zhuotao},
  booktitle={Proceedings of the Computer Vision and Pattern Recognition Conference},
  pages={14824--14834},
  year={2025}
}

@inproceedings{xu2025physaug,
  title={PhysAug: A Physical-guided and Frequency-based Data Augmentation for Single-Domain Generalized Object Detection},
  author={Xu, Xiaoran and Yang, Jiangang and Shi, Wenhui and Ding, Siyuan and Luo, Luqing and Liu, Jian},
  booktitle={Proceedings of the AAAI Conference on Artificial Intelligence},
  pages={21815--21823},
  year={2025}
}

@article{zhou2024mixstyle,
  title={Mixstyle neural networks for domain generalization and adaptation},
  author={Zhou, Kaiyang and Yang, Yongxin and Qiao, Yu and Xiang, Tao},
  journal={International Journal of Computer Vision},
  volume={132},
  number={3},
  pages={822--836},
  year={2024},
  publisher={Springer}
}

@article{wang2025yoloe,
  title={Yoloe: Real-time seeing anything},
  author={Wang, Ao and Liu, Lihao and Chen, Hui and Lin, Zijia and Han, Jungong and Ding, Guiguang},
  journal={arXiv preprint arXiv:2503.07465},
  year={2025}
}

@article{ridnik2021imagenet,
  title={Imagenet-21k pretraining for the masses},
  author={Ridnik, Tal and Ben-Baruch, Emanuel and Noy, Asaf and Zelnik-Manor, Lihi},
  journal={arXiv preprint arXiv:2104.10972},
  year={2021}
}

@inproceedings{shao2019objects365,
  title={Objects365: A large-scale, high-quality dataset for object detection},
  author={Shao, Shuai and Li, Zeming and Zhang, Tianyuan and Peng, Chao and Yu, Gang and Zhang, Xiangyu and Li, Jing and Sun, Jian},
  booktitle={Proceedings of the IEEE/CVF international conference on computer vision},
  pages={8430--8439},
  year={2019}
}
}


\end{document}